\documentclass[11pt,a4paper]{article}
\usepackage[nohyperref]{emnlp2020}
\usepackage{times}
\usepackage{latexsym}

\usepackage{microtype}

\usepackage{enumitem, graphicx, subfigure, multirow, amsfonts, amsmath, xcolor, verbatim, algorithm, algorithmicx, algpseudocode, booktabs, arydshln, times, epsfig, amssymb, multirow, capt-of}

\newcommand{\second}{\underline}

\newcommand{\eg}{\textit{e}.\textit{g}.}

\title{SSCR: Iterative Language-Based Image Editing\\via Self-Supervised Counterfactual Reasoning}
\aclfinalcopy
\def\aclpaperid{2991}

\author{Tsu-Jui Fu$^\dagger$, Xin Eric Wang$^\ddagger$, Scott T. Grafton$^\dagger$, Miguel P. Eckstein$^\dagger$, William Yang Wang$^\dagger$\\ \\
$^\dagger$UC Santa Barbara~~$^\ddagger$UC Santa Cruz \\
{\tt \small tsu-juifu@ucsb.edu, \{scott.grafton, miguel.eckstein\}@psych.ucsb.edu} \\
{\tt \small william@cs.ucsb.edu, xwang366@ucsc.edu}
}

\begin{document}
\maketitle

\begin{abstract}
Iterative Language-Based Image Editing (ILBIE) tasks follow iterative instructions to edit images step by step. Data scarcity is a significant issue for ILBIE as it is challenging to collect large-scale examples of images before and after instruction-based changes.
However, humans still accomplish these editing tasks even when presented with an unfamiliar image-instruction pair. Such ability results from counterfactual thinking and the ability to think about alternatives to events that have happened already.
In this paper, we introduce a Self-Supervised Counterfactual Reasoning (SSCR) framework that incorporates counterfactual thinking to overcome data scarcity.
SSCR allows the model to consider out-of-distribution instructions paired with previous images. With the help of cross-task consistency (CTC), we train these counterfactual instructions in a self-supervised scenario.
Extensive results show that SSCR improves the correctness of ILBIE in terms of both object identity and position, establishing a new state of the art (SOTA) on two IBLIE datasets (i-CLEVR and CoDraw). Even with only 50\% of the training data, SSCR achieves a comparable result to using complete data.
\end{abstract}

\section{Introduction}





Digital design tools like Illustrator or Photoshop are widely used nowadays. Though having considerable user demand, they require prior knowledge and multiple steps to use successfully. These applications would significantly improve the accessibility if they can automatically perform corresponding editing actions based on the language instructions given by users for each step.



\begin{figure}[t]
    \centering
    
    \includegraphics[width=\linewidth]{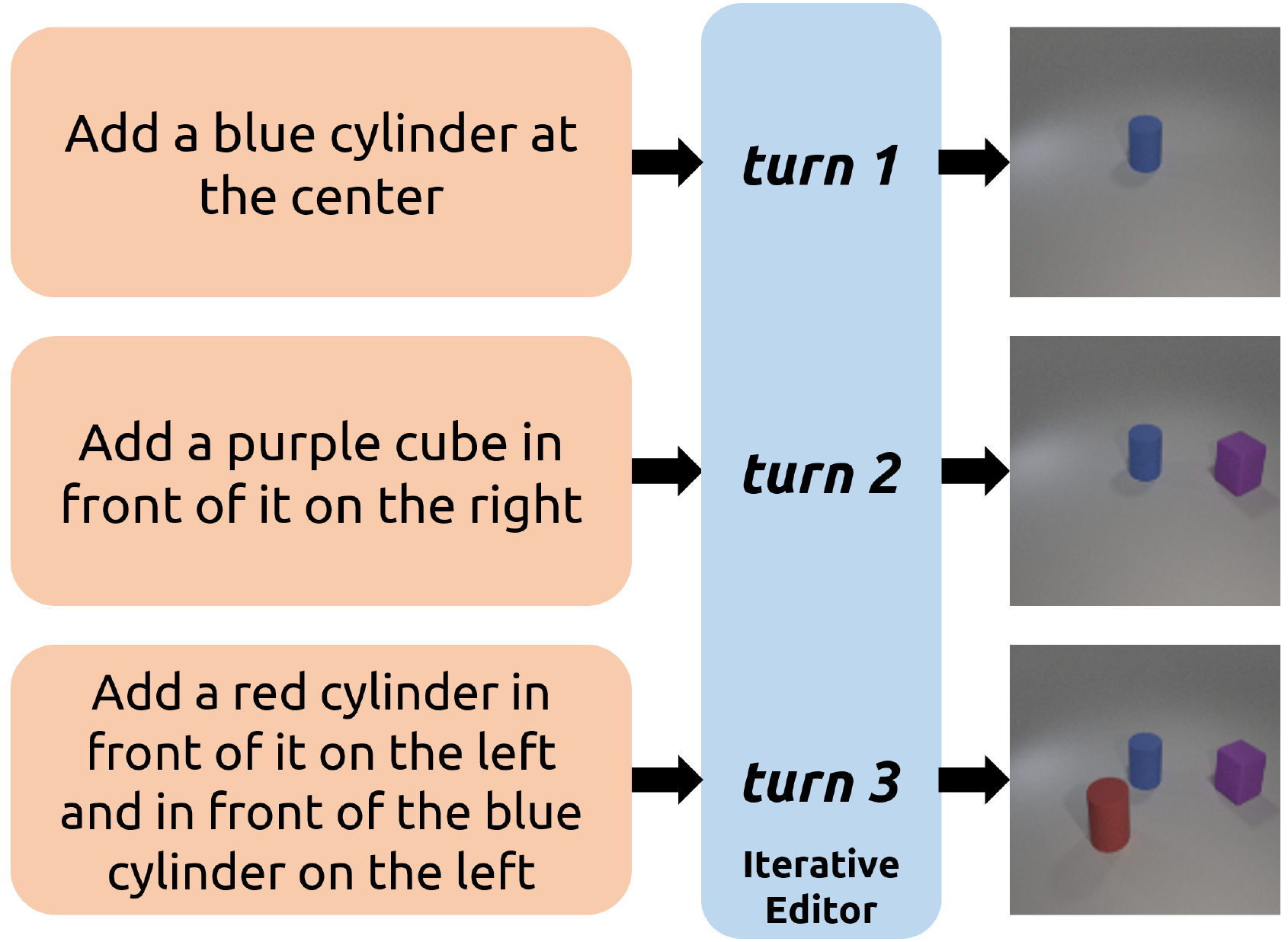}
     \vspace{-3ex}
    \caption{An example of the iterative language-based image editing (ILBIE) task. During each turn, the model edits the image from the previous turn based on the current instruction. Eventually, a desired image is accomplished after iterative editing. Note that the generation is at the pixel level.}
    \label{fig:task}
    \vspace{-3ex}
\end{figure}

Iterative language-based image editing (ILBIE) task follows iterative instructions to edit images step by step, as illustrated in Fig.~\ref{fig:task}. To accomplish ILBIE, models are required not only to modify images but also to understand the visual differences between the previous and resulting image, based on the given instructions. One of the primary limitations of ILBIE is data scarcity. Since collecting large-scale previous-resulting images with instructions is difficult, it makes learning the association between vision and language challenging.



A GAN-based \cite{goodfellow2015gan} model, GeNeVA~\cite{nouby2019geneva}, is proposed to perform ILBIE, where a conditional generator serves as the image editor, and a discriminator provides the training loss by discriminating a resulting image.
Though it yields promising results, GeNeVA neglects the data scarcity issue of ILBIE.
As a binary classifier, the discriminator easily suffers from data scarcity given the shortage of ground-truth instruction-and-resulting-image pairs and thus limits the model's generalization ability to new instructions.



Despite lacking prior experiences with images or instructions, humans can still accomplish editing under unfamiliar image-instruction pairs.
For example, for a given instruction ``add a purple cube in front of the blue cylinder", humans can think about the resulting image if the instruction changed to ``adding a blue square" or ``on the right of". This process is known as counterfactual thinking \cite{roese1997ctf}, which allows humans to operate in data-scarce scenarios by considering alternative instructions from the seen examples.
 
In this paper, we introduce self-supervised counterfactual reasoning (SSCR) that incorporates counterfactual thinking to deal with the data scarcity issue. SSCR allows the model to think about expected, resulting images under unseen instructions. Since there are no ground-truth resulting images, we propose cross-task consistency (CTC), which adopts an iterative explainer to reconstruct the instruction of each step. With CTC, we can supply detailed token-level training loss (e.g., wrong objects or wrong positions), which is better than only using the discriminator, and consider these counterfactual instructions in a self-supervised scenario.


The experimental results on i-CLEVR~\cite{nouby2019geneva} and CoDraw~\cite{kim2019codraw} show that our SSCR can improve the correctness of the ILBIE task in both aspects of object identity and position. In summary, our contributions are three-fold:
\begin{itemize}[noitemsep, topsep=2pt]
    \item We introduce SSCR that incorporates counterfactual thinking into the ILBIE task to deal with the data scarcity issue.
    \item The proposed cross-task consistency (CTC) and counterfactual reasoning methods help train the generator better, improve the generalizability, and achieve the SOTA results on i-CLEVR and CoDraw.
    \item Extensive ablation studies show that SSCR is effective even with only partial training data.
\end{itemize}

\section{Related Work}
\noindent\textbf{Text-to-Image} (T2I) 
generates an image that matches the given instruction. T2I is challenging yet important that has a vast potential in practical applications like art generation or automatic design \cite{nguyen2017langevin,reed2017pixcelcnn, tan2019t2s}. With the success of a generative adversarial network \cite{goodfellow2015gan} on the image generation task, several works \cite{reed2016t2i-gan,zhang2017stack-gan, xu2018att-gan} introduce different GAN-based models to synthesize an image from a text description. Unlike T2I, we focus on image editing, where a model needs to understand the visual differences between two images rather than generating an image from scratch.


\noindent\textbf{Language-based Image Editing} 
(LBIE) tasks a model to edit an image based on the guided text description. PixelTone \cite{laput2013pixel-tone} and Image Spirit \cite{cheng2013image-spirit} are both rule-based, which accept only pre-defined instructions and semantic labels that can significantly decrease the practicality of LBIE. Some studies \cite{chen2018lbie-ram,shinagawa2017edit-mnist} adopt the conditional GAN model to attend on the instruction and perform LBIE as image colorization. However, image colorization is not truly an editing task since it only supports fixed object templates, and the objects and the scene of the image remain the same after editing. In contrast, the editing processes of Photoshop or Illustrator are not accomplished in a single pass. GeNeVA \cite{nouby2019geneva} proposes an iterative GAN-based generator to accomplish iterative language-based image editing (ILBIE) but neglects the data scarcity issue. 

\noindent\textbf{Counterfactual Thinking}
~\cite{roese1997ctf} is a concept that refers to the human propensity to consider possible alternatives to events that have happened already. People can consider different outcomes from a wide range of conditions and engage in causal reasoning by asking questions like ``What if ...?" or ``If I had only...." Previous works \cite{kusner2017ctf-fair,garg2019ctf-tc} have shown how counterfactual fairness improves the robustness of the model and makes it more explainable. Furthermore, counterfactual thinking has also been applied to augment training targets \cite{zmigrod2019ctf-da,fu2020aps}. In this paper, we incorporate counterfactual thinking into that ILBIE task that considers counterfactual instructions to deal with the data scarcity issue and improve the generalizability.

\begin{figure*}[t]
    \centering
    
    \includegraphics[width=\linewidth]{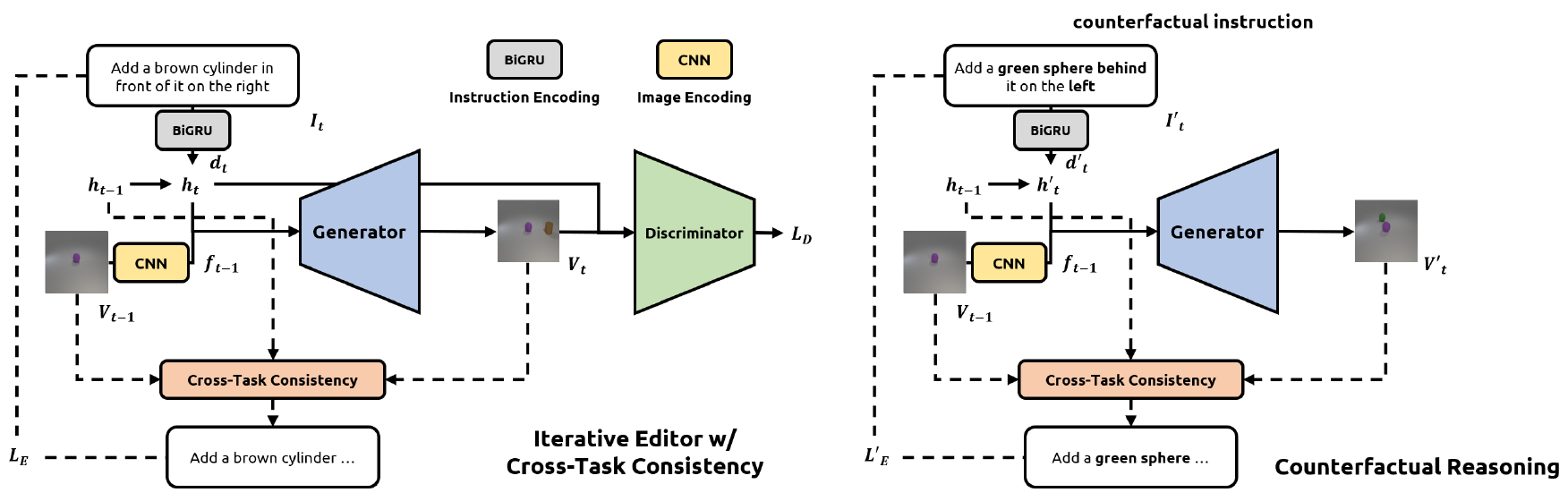}
    \vspace{-3ex}
    \caption{An overview of our self-supervised counterfactual reasoning (SSCR). The iterative editor modifies an image based on current instruction and editing history. Counterfactual reasoning allows the model to think about various counterfactual instructions that can improve the generalizability and deal with data scarcity. Since there are no ground-truth images, we propose cross-task consistency (CTC) to not only provide explicit training signal but also train these counterfactual instructions self-supervisedly.}
    \label{fig:overview}
    \vspace{-3ex}
\end{figure*}

\section{Self-Supervised Counterfactual Reasoning (SSCR)}
\vspace{-1ex}

\subsection{Task Definition}
During each turn $t$, an editor edits the image from the previous turn $V_{t-1}$ into the current turn $V_t$ based on instruction $I_t$. After a final turn $T$, we get the predicted final image $V_T$ and evaluate the outcome with the ground truth resulting image $O_T$. Note that the editing process is at a pixel level where the model has to generate each pixel of the image:
\begin{equation}
\begin{split}
    V_t &= \text{Editor}(V_{t-1}, I_t), \\
    \text{eval} &= \text{Compare}(V_T, O_T).
\end{split}
\end{equation}

\subsection{Overview}
To overcome data scarcity, we introduce self-supervised counterfactual reasoning (SSCR). The overall framework is illustrated in Fig,~\ref{fig:overview}. The iterative editor is a conditional generator that modifies an image based on current instruction and editing history.

Counterfactual reasoning allows the model to think about the expected, resulting images under various counterfactual instructions. In this way, the editor can consider more diverse instructions than the original data to improve the generalizability, even if under data scarcity. Since there are no ground-truth resulting images for these counterfactual instructions, we propose cross-task consistency (CTC). CTC adopts an iterative explainer to reconstruct the given instruction of each editing step. With the help of this cross-task matching, we can not only provide a detailed token-level training signal to train the editor better but also supply training loss for counterfactual reasoning in a self-supervised scenario.

\subsection{Iterative Editor}
Similar to GeNeVA \cite{nouby2019geneva}, the iterative editor is a GAN-based architecture that contains a conditional generator $G$ and a discriminator $D$. We first apply a bidirectional GRU \cite{chung2014gru} to encode the instruction $I_t$ as $d_t$ for each turn $t$. And another GRU is used to encode the history of instructions $h_t$ as following:
\begin{equation}
    h_t = \text{GRU}(d_t, h_{t-1}).
\end{equation}

Then, to perform the editing for turn $t$, we adopt a convolutional neural network (CNN) \cite{miyato2018res-block} to extract image features $f_{t-1}$ from the previous image $V_{t-1}$, concatenate with the instruction history $h_t$, and feed into $G$ to predict the resulting image $V_{t}$:
\begin{equation}
    V_t = G([f_{t-1}, h_t]).
\end{equation}
After all iterations, there is the final image $V_{T}$ after the final turn $T$.

For each turn, $D$ provides a binary training signal by discriminating a resulting image that is generated from either $G$ or the ground-truth data according to the instruction history $h_t$:
\begin{equation} \label{eq:lg}
    L_G = \sum_{t=1}^{T} \mathbb{E}_{V_t \sim \mathcal{P}_{G_t}}[\log(D([V_t, h_t]))],
\end{equation}
where $L_G$ is the binary loss from $D$.

For training $D$, similar to T2I \cite{reed2016t2i-gan}, we add additional [real image, wrong instruction] pairs as false examples:
\begin{equation} \label{eq:ld}
    L_D = \sum_{t=1}^{T} L_{D_{\text{real}_t}}+\frac{1}{2}(L_{D_{\text{false}_t}}+L_{D_{\text{wrong}_t}}),
\end{equation}
where
\begin{equation}
\begin{split}
    L_{D_{\text{real}_t}} &= \mathbb{E}_{O_t \sim \mathcal{P}_\text{data}}[\log(D([O_t, h_t]))], \\
    L_{D_{\text{false}_t}} &= \mathbb{E}_{V_t \sim \mathcal{P}_{G_t}}[\log(1-D([V_t, h_t]))], \\
    L_{D_{\text{wrong}_t}} &= \mathbb{E}_{O_t \sim \mathcal{P}_\text{data}}[\log(1-D([O_t, h'_t]))],
\end{split}
\end{equation}
with ground-truth data distribution $\mathcal{P}_\text{data}$ and $h'_t$ being the wrong instruction history by randomly selecting another instruction.

Then $G$ and $D$ are optimized through an alternating minmax game:
\begin{equation}
\begin{split}
    \max_{G} & \min_{D} L_G+L_D.
\end{split}
\end{equation}

\subsection{Cross-Task Consistency (CTC)}
Though we can train the iterative editor for ILBIE, $D$ only supports a binary training loss, which is not explicit enough to express the complex association between the visual difference and the text description.
To supply a more explicit training loss, we propose cross-task consistency (CTC).
Despite being image generation, we consider instruction generation, which explains the visual difference between previous-resulting image pairs, to do reasoning for the editing process in a cross-task scenario.
During CTC, an iterative explainer provides a token-level training signal that encourages the matching between the predicted image and the original instruction.

\begin{figure}[t]
    \centering
    
    \includegraphics[width=\linewidth]{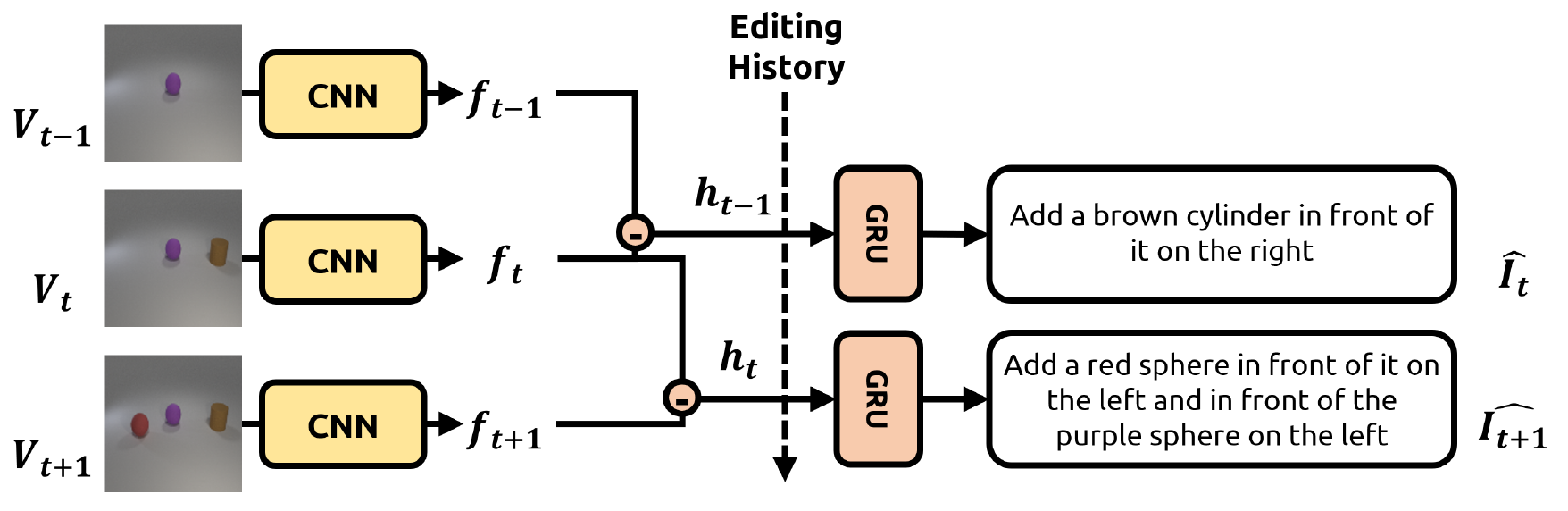}
    \vspace{-3ex}
    \caption{The architecture of our iterative explainer. We consider the previous-resulting image pair and the encoded instruction history as input to reconstruct the editing instruction by an attention-based GRU decoder.}
    \label{fig:explainer}
    \vspace{-3ex}
\end{figure}

\paragraph{Iterative Explainer}
Our iterative exaplainer $E$ is an instruction decoder which considers previous-resulting image pair and the instruction history as input, as shown in Fig.~\ref{fig:explainer}: 
\begin{equation}
    \hat{I_t} = E(V_t, V_{t-1}, h_{t-1}).
\end{equation}

Similar to the iterative editor, we apply CNN to extract visual feature $f$ for both previous and predicted resulting image:
\begin{equation}
    f_{t-1} = \text{CNN}(V_{t-1}), f_t = \text{CNN}(V_t).
\end{equation}

Then, a GRU serves as an attention-based language decoder \cite{xu2015att} which reconstructs the instruction $\hat{I_t}$ according to the feature difference and instruction history $h_{t-1}$ of previous turn:
\vspace{-2ex}
\begin{equation}
\begin{split}
    g_0 = [f_d, h_{t-1}],\\
    \hat{w_i}, g_i = \text{GRU}(w_{i-1}, g_{i-1}), \\
    \hat{I_t} = \{\hat{w_1}, \hat{w_2}, ..., \hat{w_L}\},
\end{split}
\end{equation}
where $f_d = f_t-f_{t-1}$ represents the visual difference by subtracting previous and result feature, $g_i$ is the decoding history, and $\hat{w_i}$ is the predicted word token of the instruction. All $w_i$ are combined as the reconstruction where $L$ is the length of the instruction. The iterative explainer considers not only the visual difference but also instruction history so that we can reconstruct the instruction, which explains the editing of the resulting image following by the editing history.

Finally, we provide an explicit token-level training signal $L_E$ by computing the teacher-forcing loss \cite{williams1989tf} between the original instruction $I_t$ and the reconstructed one $\hat{I_t}$:
\begin{equation} \label{eq:le}
    L_E = \sum_{i=1}^{L}\text{CELoss}(\hat{w_i}, w_i), 
\end{equation}
where $w_i$ is the $i$th token of $I_t$ and CELoss means the cross-entropy loss.

By minimizing $L_E$, $G$ learns to match the original instruction with this cross-task consistency. Different from $L_G$, which only supplies binary but vague loss, $L_E$ provides token-level loss about the information of the wrong object or wrong position (by comparing $\hat{w_i}$ with $w_i$) that can train $G$ better for each editing turn. In the experiments, $E$ is pre-trained by the ground-truth image pairs and is fixed during the following training.

\begin{table}[t]
    \small
    \centering
    
    \begin{tabular}{ccc}
        \toprule
        Dataset & Token Type & Example \\
        \midrule
        \multirow{3}{*}{i-CLEVR} & color & blue, purple \\
        & object & cylinder, cube \\
        & relation & at the center, in front of \\
        \midrule
        \multirow{3}{*}{CoDraw} & size & small, meidum \\
        & object & sun, boy \\
        & relation & in the middle, on the left \\
        \bottomrule
    \end{tabular}
    \vspace{-1ex}
    \caption{Types of token on i-CLEVR and CoDraw.}
    \label{table:token}
    
    \vspace{-3ex}
\end{table}

\subsection{Counterfactual Reasoning}
We assume that $\mathcal{U}$ is the available training data. Because of the practical challenge of collecting large-scale previous-resulting images with instructions, $\mathcal{U}$ suffers from data scarcity. To deal with this issue, we propose counterfactual reasoning to allow the model to consider various instructions out of the distribution of $\mathcal{U}$. For instance, an instruction $I' \sim \mathcal{U}'$ from the intervention data $\mathcal{U}'$ replaces the original instruction, and we edit the image based on the counterfactual instruction $I'$. 

\paragraph{Instruction Intervention}
To get the intervention data $U'$ that provides diverse instructions, we do interventions on the original instructions $I$:
\begin{equation}
\begin{split}
    I, O &= \mathcal{U}, \\
    I' &= \text{intervention}(I), \\
    \mathcal{U}' &= \{I', O\},
\end{split}
\end{equation}
where $O$ is the image in the original $\mathcal{U}$.

First, we apply NLTK \cite{pbird2004nltk} to parse out \textit{tokens} in the original $I$. The types of \textit{token} on i-CLEVR and CoDraw are shown in Table~\ref{table:token}. 

We then replace these \textit{tokens} with randomly sampled \textit{tokens} of the same type to get the counterfactual $I'$. Finally, $I'$ combines with the original image $G$ as the intervention data $\mathcal{U}'$. Our experiments show that this simple yet effective intervention makes the training data more diverse and deals with data scarcity during our counterfactual reasoning.

For each turn $t$, with $I'_t$ from $U'$, we predict the counterfactual resulting image $V'_t$:
\begin{equation}
    V'_t = G([f_{t-1}, h'_t]),
\end{equation}
where $h'_t$ is the counterfactual instruction history encoded from $I'$.

Since there is no ground-truth image for the counterfactual instruction $I'_t$, we adopt the iterative explainer $E$ to provide counterfactual training loss $L'_E$ in a self-supervised scenario:
\begin{equation}
\begin{split} \label{eq:cf}
    \hat{I'_t} &= E(V'_t, V_{t-1}, h_{t-1}), \\
    L'_E &= \sum_{i=1}^{L}\text{CELoss}(\hat{w'_i}, w'_i), 
\end{split}
\end{equation}
where $\hat{w'_i}$ and $w'_i$ are the $i$th word token.

\begin{algorithm}[t]
    \begin{algorithmic}[1]
        \small
        
        \State $G$: the generator model
        \State $D$: the discriminator model
        \State $H$: the instruction history encoder
        \State $E$: our iterative explainer in CTC
        \State $\mathcal{U}$: the original training set
        \\
        \State Pre-train RE with $\mathcal{U}$
        \State Initialize G, D
        \\
        \While{TRAIN\_EDITOR}
            \For{$t \gets 1$ to $T$}
                \State $I_t$, $O_t$ $\gets$ $\mathcal{U}$
                \State $h_t$ $\gets$ $H(h_{t-1}, I_t)$
                \State $V_t$ $\gets$ $G(h_t, O_{t-1})$ \Comment{teacher-forcing training}
                \State $\hat{I_t}$ $\gets$ $E(V_t, O_{t-1}, h_{t-1})$
                \\
                \State $L_G$ $\gets$ binary loss from $D$ \Comment{Eq.~\ref{eq:lg}}
                \State $L_E$ $\gets$ explicit loss from $E$ \Comment{Eq.~\ref{eq:le}}
                \State Update $G$ by maximizing $L_G-L_E$
                \\
                \State $L_D$ $\gets$ loss for $D$ \Comment{Eq.~\ref{eq:ld}}
                \State Update $D$ by minimizing $L_D$
            \EndFor
        \EndWhile
    \end{algorithmic}

    \caption{Iterative Editor with CTC}
    \label{algo:ctc}
\end{algorithm}

\begin{algorithm}[t]
    \begin{algorithmic}[1]
        \small
        
        \While{COUNTERFACTUAL\_REASONING}
            \For{$t \gets 1$ to $T$}
                \State $\mathcal{U}'$ $\gets$ intervention($\mathcal{U}$)
                \State \_, $O_t$ $\gets$ $\mathcal{U}$
                \State $I'_t$, \_ $\gets$ $\mathcal{U}'$ \Comment{counterfactual instruction}
                \\
                \State $h_t$ $\gets$ $H(h_{t-1}, I_t)$ \Comment{real history}
                \State $h'_t$ $\gets$ $H(h_{t-1}, I'_t)$ \Comment{counterfactual history}
                \State $V'_t$ $\gets$ $G({h'}_t, O_{t-1})$ \Comment{counterfactual editing}
                \State $\hat{I'_t}$ $\gets$ $E({V'}_t, O_{t-1}, h_{t-1})$
                \\
                \State $L'_E$ = counterfactual loss from $E$ \Comment{Eq.~\ref{eq:cf}}
                \State Update $G$ by minimizing ${L'}_E$
            \EndFor
        \EndWhile
    \end{algorithmic}

    \caption{Counterfactual Reasoning}
    \label{algo:sscr}
\end{algorithm}

By minimizing $L'_E$, the model has an opportunity to access $U'$, which is different from the original training data. With the help of our iterative explainer, SSCR improves the generalizability by reasoning diverse counterfactual instructions $I'$ even if under data scarcity.

\begin{table*}
    \centering
    \small
    \begin{tabular}{lccccccccc}
        \toprule
        ~ & \multicolumn{4}{c}{\textbf{i-CLEVR}} & ~ & \multicolumn{4}{c}{\textbf{CoDraw}} \\
        \cmidrule{2-5} \cmidrule{7-10}
        Method & Precision & Recall & F1 & RelSim & ~ & Precision & Recall & F1 & RelSim \\
        \midrule
        GeNeVA & 71.01 & 42.61 & 53.26 & 30.66 & ~ & 54.38 & 54.42 & 54.40 & \second{38.93}  \\ 
        w/ CTC only & \second{72.24} & \second{45.51} & \second{55.84} & \second{33.67} & ~ & \second{57.69} & \second{55.60} & \second{56.62} & 38.68 \\
        w/ SSCR & \textbf{73.75} & \textbf{46.39} & \textbf{56.96} & \textbf{34.54} & ~ & \textbf{58.17} & \textbf{56.61} & \textbf{57.38} & \textbf{39.11} \\
        \bottomrule
    \end{tabular}
    \vspace{-1ex}
    \caption{The testing results of the baseline (GeNeVA\footnotemark[1]), with only cross-task consistency (CTC only), and with whole self-supervised counterfactual reasoning (SSCR) for both i-CLEVR and CoDraw.}
    \label{table:result}
    
    \vspace{-3ex}
\end{table*}

\subsection{Learning of SSCR}
Alg.~\ref{algo:ctc} presents the learning process of training the iterative editor with CTC.
Since ILBIE is also a sequential generation process, we apply the widely used teacher-forcing where we feed in the ground-truth resulting image ($O_{t-1}$) from the previous turn instead of our predicted one ($V_{t-1}$) to make the training more robust.
When training the iterative editor, for each turn $t$, we adopt $G$ to perform image editing. We maximize the binary loss from $D$ ($L_G$) with minimizing the explicit token-level loss from $E$ ($L_E$) to train $G$. We also update $D$ by minimize $L_D$:
\begin{equation}
\begin{split}
    \max_{G} & \min_{D} L_G+L_D-L_E.
\end{split}
\end{equation}

During counterfactual reasoning, as shown in Alg.~\ref{algo:sscr}, we first perform an intervention on $\mathcal{U}$ to get the counterfactual instructions ($I'$). Then, we edit the image based on $I'$. Since there is no ground-truth resulting image for the counterfactual editing, we adopt CTC to compute the cycle-consistency loss ($L'_E$) self-supervisedly. Similar to the iterative editor, we also apply teacher-forcing training (feeding in $O_{t-1}$ and $h_{t-1}$) to further update $G$.
In this way, $G$ can improve the generalizability by considering the counterfactual $U'$, which is more diverse than $U$.

\section{Experiments}
\vspace{-1ex}

\subsection{Experimental Setup}
\paragraph{Datasets}
We evaluate our counterfactual framework on two ILBIE datasets, i-CLEVR \cite{nouby2019geneva} and CoDraw \cite{kim2019codraw}. Each example in i-CLEVR consists of a sequence of 5 (image, instruction) pairs. The instruction describes where the object should be placed relative to existing objects. In i-CLEVR, there are 6,000, 2,000, and 2,000 examples for training, validation, and testing, respectively.

CoDraw is a more difficult art-like dataset of children playing in a park. There are 58 objects and children with different poses. We use the same split as in CoDraw where 7,988, 1,002, and 1,002 are for training, validation, and testing.

\paragraph{Evaluation Metrics}
Standard metrics like Inception Score \cite{salimans2016is} or Frechet Inception Distance (FID) \cite{heusel2017fid} cannot detect whether the editing is correct based on the instruction \cite{nouby2019geneva}. Following GeNeVA \cite{nouby2019geneva}, we adopt F1 and RelSim to evaluate the editing result.

The F1 score is based on a pre-trained object detector \cite{szegedy2016v3} (\textasciitilde 99\% accuracy on both i-CLEVR and CoDraw), which detects the objects in the predicted images that meet the ground-truth resulting images. To evaluate not only object type but also object position, we build the scene graph according to the object detector. The edges are given by the left-right and front-back relations between the vertices (objects). Then, RelSim determines how many of the ground-truth relations are in the predicted images:
\begin{equation}
    \text{RelSim}(E_\text{gt}, E_\text{pd}) = \text{recall} \times \frac{|E_\text{pd} \cap E_\text{gt}|}{|E_\text{gt}|},
\end{equation}
where $E_\text{gt}$ and $E_\text{pd}$ are relational edges for ground-truth resulting images and predicted image. Note that we only evaluate the final predicted image of each example for both F1 and RelSim.

\paragraph{Baseline}
We use the SOTA model GeNeVA\footnote{We reproduce the result for GeNeVA by their official GitHub repo ({https://github.com/Maluuba/GeNeVA}). We apply the default hyperparameters as them, and issue \#2 can support that the results are comparable.} as our baseline: it shares the same model architecture as our iterative editor and is trained with the GAN objective but without the cross-task consistency (CTC) and our counterfactual reasoning. 

\paragraph{Implementation Detail}
We apply the ResBlocks \cite{miyato2018res-block} into $G$ and $D$ where the visual feature size is 1024. For our $E$, we add self-attention \cite{zhang2019sa-gan} for the concatenation of the visual difference and the encoded instruction history. We adopt Adam \cite{kingma2015adam} to optimize the iterative editor with learning rate 1e-4 for $L_G$ and $L_E$, 4e-4 for $L_D$. The learning rate of $L'_E$ during the counterfactual reasoning is 5e-5.

\begin{figure*}[t]
    \centering
    
    \begin{minipage}{.59\linewidth}
        \centering
        \includegraphics[width=\linewidth]{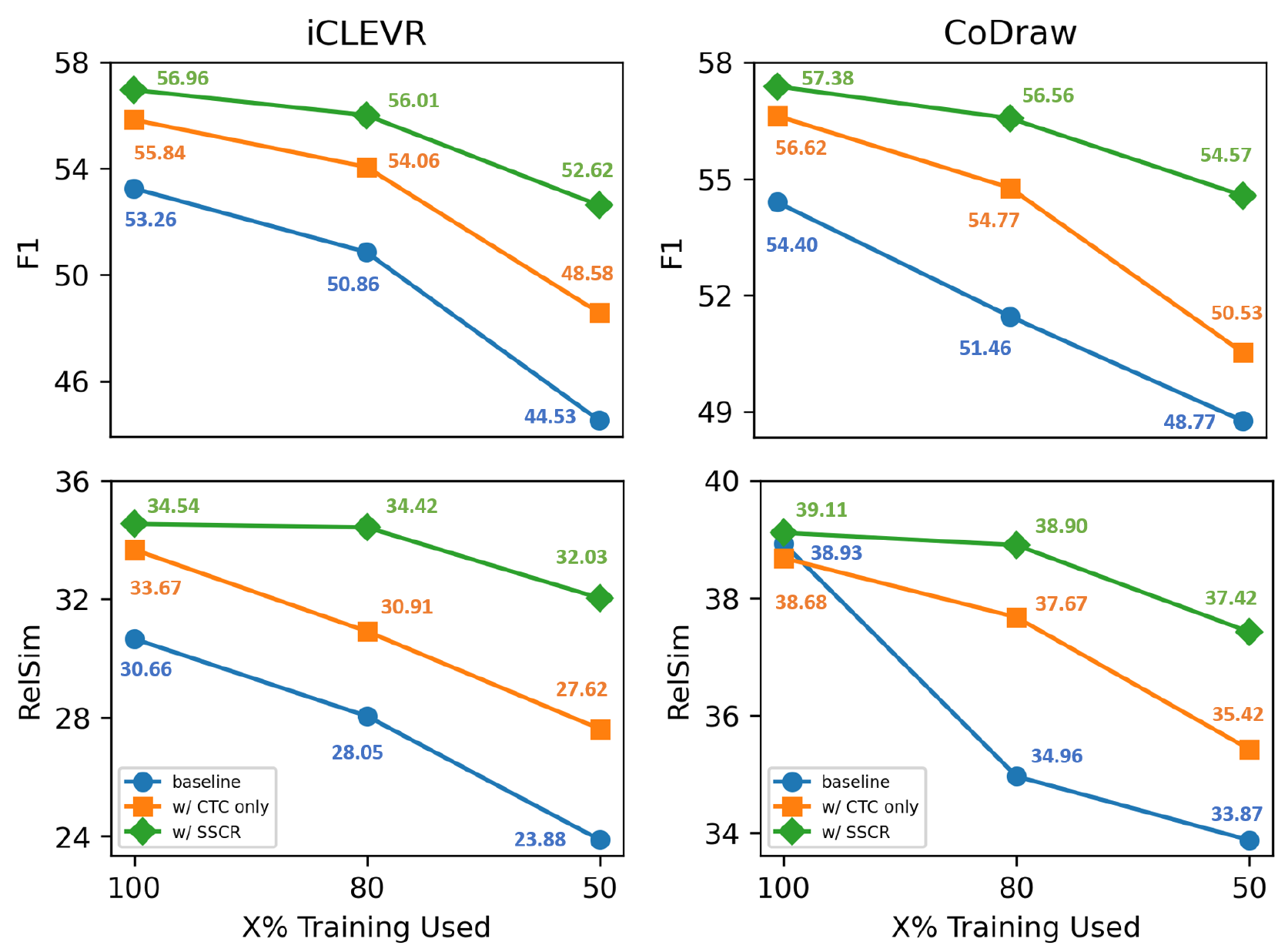}
        \vspace{-3ex}
        \captionof{figure}{Result comparison among baseline, with only cross-task consistency (CTC only), and with whole self-supervised counterfactual reasoning (SSCR) under different ratios of training data. Note that the iterative explainer is also pre-trained using the same available data for each result.}
        \label{fig:ratio}
        \vspace{-3ex}
    \end{minipage}~~~
    \begin{minipage}{.39\linewidth}
        \centering
        \includegraphics[width=.8\linewidth]{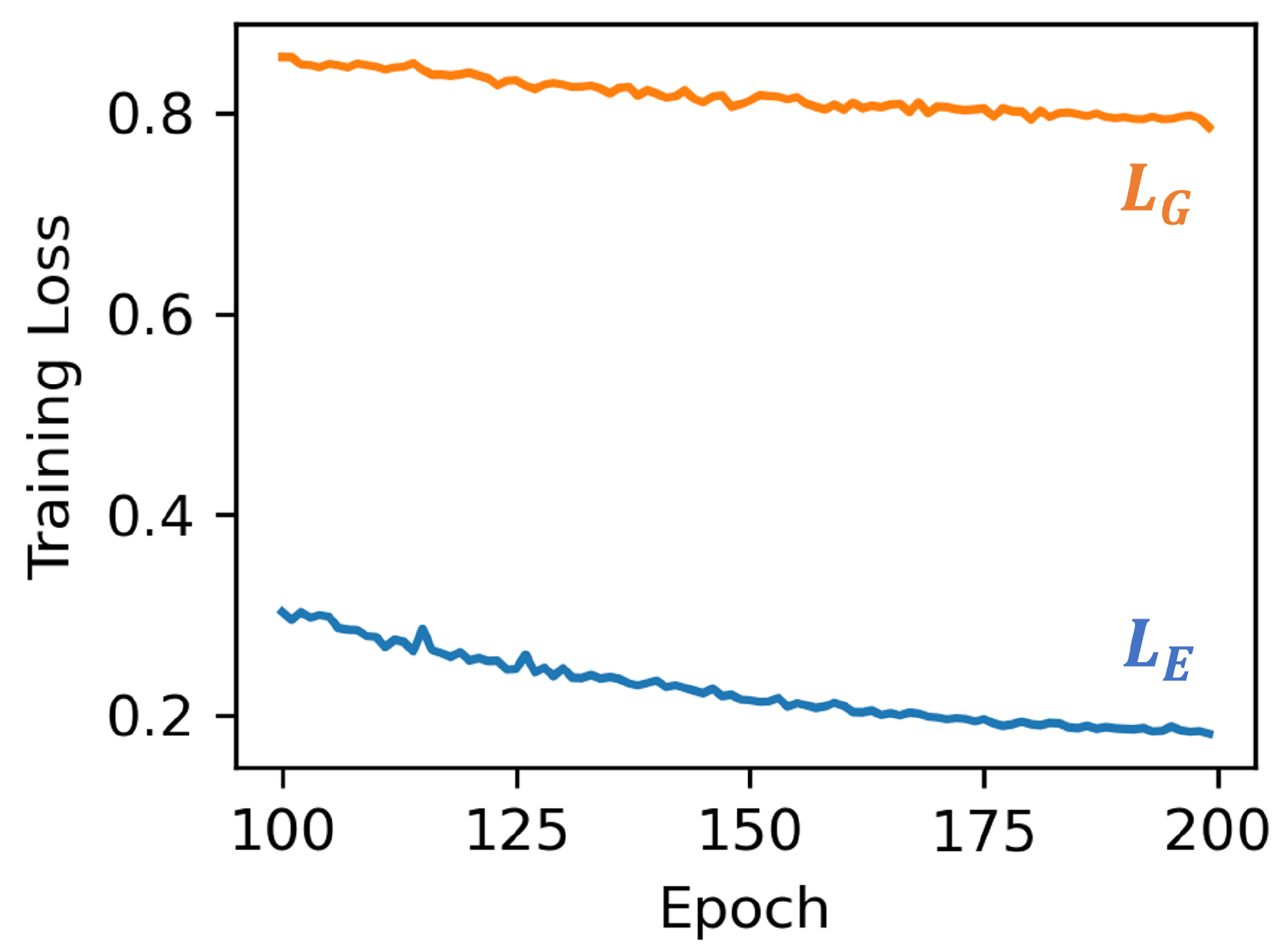}
        \vspace{-1ex}
        \captionof{figure}{The learning curve of training loss provided from the discriminator ($L_G$) and our iterative explainer ($L_E$) on i-CLEVR.}
        \label{fig:loss-vs}
        
        \vspace{2ex}
        
        \small
        \begin{tabular}{ccc}
            \toprule
            X\% Training Used & PPL & BLEU \\
            \midrule
            100\% & 0.1073 & 50.236 \\
            ~~80\% & 0.1295 & 48.873 \\
            ~~50\% & 0.1163 & 48.763 \\
            \bottomrule
        \end{tabular}
        \vspace{-1ex}
        \captionof{table}{The PPL and BLEU of our iterative explainer with different ratios of training data on i-CLEVR.}
        
        \label{table:explainer}
        \vspace{-3ex}
    \end{minipage}
\end{figure*}

\subsection{Quantitative Results}
Table~\ref{table:result} presents the testing F1 and RelSim results. First, with our cross-task consistency (CTC only), which provides a more explicit training signal, we can improve the baseline on the i-CLEVR dataset in terms of all metrics. In particular, CTC improves 1.2\% on precision, 2.9\% on recall, and 2.6\% on F1. Additionally, for whole self-supervised counterfactual reasoning (SSCR), which allows the model to consider out-of-distribution instructions, it brings more improvements and achieves new SOTA results, \eg, 56.9\% on F1 and 34.5\% on RelSim.

Similar trends can be found on CoDraw. Since the instructions under CoDraw are more complex, the improvement of relation correctness (RelSim) is not as high as i-CLEVR. But for object correctness, CTC still improves baseline with 2.2\% F1, and SSCR further achieves the new SOTA on all metrics, \eg, 57.4\% F1 and 39.1\% RelSim.

\subsection{Ablation Study}
\paragraph{Under Data Scarcity}
To examine the framework's effectiveness under the data scarcity scenario, we compare models trained using 100\%, 80\%, and 50\% data. Note that our $E$ is also pre-trained using the same 100\%, 80\%,  and 50\% data.
The results are shown in Fig.~\ref{fig:ratio}.

We can observe that on both i-CLEVR and CoDraw datasets, the baseline performance drops drastically as the training data decreases, and our SSCR consistently outperforms the baseline. More importantly, the baseline severely suffers from the data scarcity issue, while SSCR is relatively resilient to data decrease and only drops 4.34 F1 score and 2.51 RelSim score (\textit{vs.} 8.73 and 6.78 reduced by the baseline) on iCLEVR when there is only 50\% data. Similar results can be observed on CoDraw. Furthermore, comparing SSCR with 50\% data and the baseline with 100\%, we can notice that our method can achieve comparable results to the baseline with only half the data used for training. Therefore, incorporating counterfactual thinking to explore out-of-distribution instructions indeed makes the model better capable of generalization and avoiding performances drops from data scarcity.

Table~\ref{table:explainer} presents the performance of our iterative explainer $E$ with different ratios of training examples. Perplexity (PPL) and BLEU \cite{papineni2002bleu} are calculated between the reconstructed instructions and the original ones. We can see that the PPL and BLEU under 50\% are similar to 100\%. It shows that $E$ still supplies meaningful training loss for SSCR even if only using 50\% data.

\begin{table}
    \centering
    \small
    
    \begin{tabular}{lccccc}
        \toprule
        ~ & \multicolumn{2}{c}{100\%} & ~ & \multicolumn{2}{c}{50\%} \\
        \cmidrule{2-3} \cmidrule{5-6}
        Method & F1 & RelSim & ~ & F1 & RelSim \\ 
        \midrule
        GeNeVA & 53.26 & 30.66 & ~ & 44.53 & 23.88 \\
        w/ SSCR (D) & 54.05 & 30.87 & ~ & 43.31 & 22.99 \\
        w/ SSCR (E) & \textbf{56.95} & \textbf{34.54} & ~ & \textbf{52.62} & \textbf{32.03} \\
        \bottomrule
    \end{tabular}
    \vspace{-1ex}
    \caption{Results of discriminator ($D$) or iterative explainer ($E$) used for the counterfactual reasoning (SSCR) on i-CLEVR.}
    \label{table:sscr-vs}
    \vspace{-1ex}
\end{table}

\begin{table}
    \centering
    \small
    
    \begin{tabular}{lcc}
        \toprule
        Method & F1 & RelSim \\ 
        \midrule
        GeNeVA & 42.23 & 23.70 \\
        w/ CTC Only & \second{43.91} & \second{25.26} \\
        w/ SSCR & \textbf{48.30} & \textbf{29.09} \\
        \bottomrule
    \end{tabular}
    \vspace{-1ex}
    \caption{Results of zero-shot generalization.}
    
    \label{table:zero-shot}
    \vspace{-3ex}
\end{table}

\begin{figure*}[t]
    \centering
    
    \includegraphics[width=0.8\linewidth]{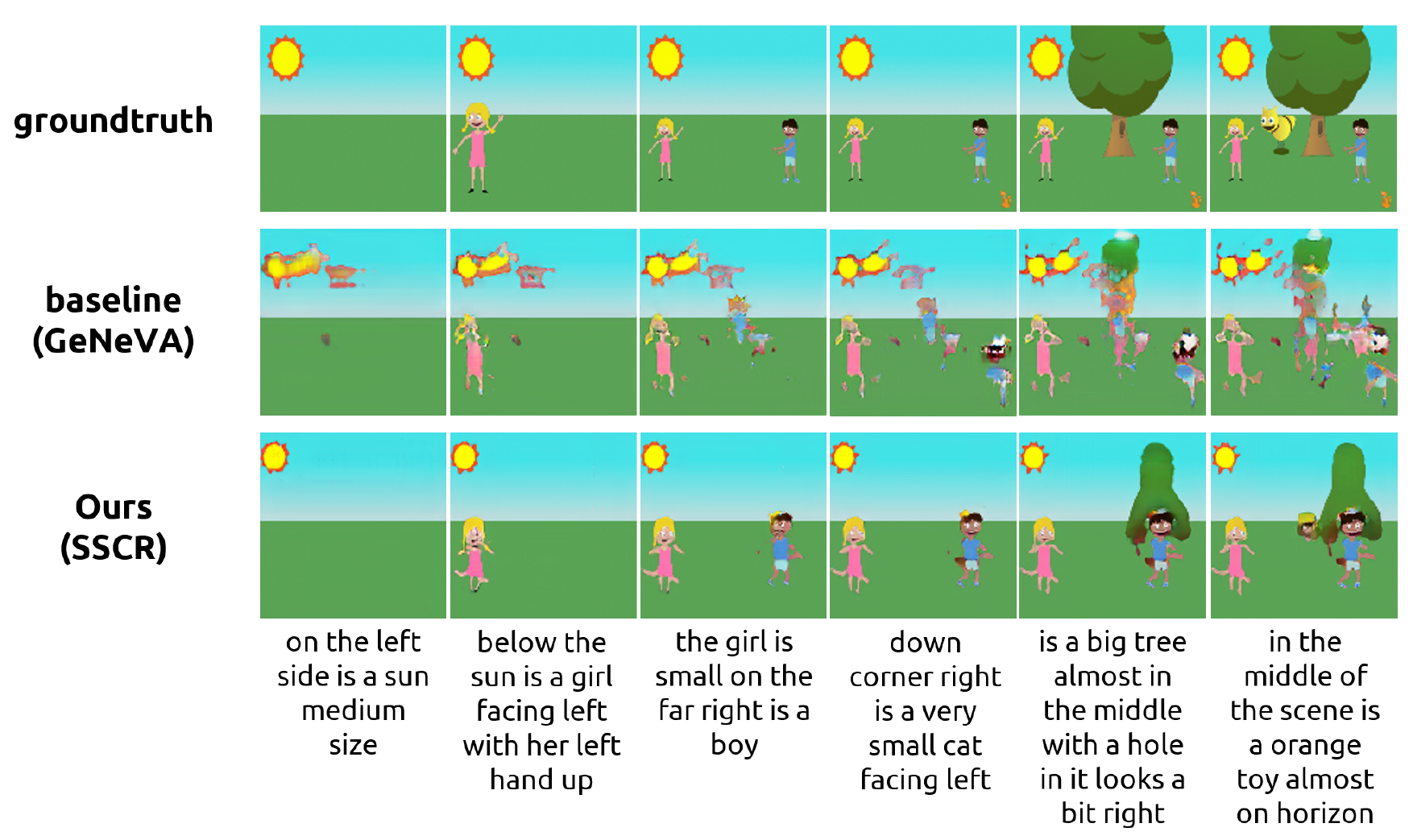}
    \vspace{-1ex}
    \caption{Visulaization example of baseline (GeNeVA) and our SSCR on CoDraw.}
    
    \label{fig:case}
    \vspace{-1ex}
\end{figure*}

\paragraph{Zero-shot Generalization}
To further demonstrate the effectiveness of SSCR under severe data scarcity, we conduct a zero-shot experiment for the i-CLEVR dataset. The zero-shot setting is as following. There are 3 shapes (\textit{cube}, \textit{sphere}, \textit{cylinder}) and 8 colors (\textit{gray}, \textit{red}, \textit{blue}, \textit{green}, \textit{brown}, \textit{purple}, \textit{cyan}, \textit{yellow}), which lead to 24 different objects on the i-CLEVR dataset. We remove examples containing “\textit{gray cube}, \textit{red cube}, \textit{green sphere}, or \textit{purple cylinder}” in the training set but still evaluate the full testing set with all kinds of objects.

The result is shown in Table ~\ref{table:zero-shot}. Since there is no example like “\textit{gray cube}” in the training set, CTC can only consider those seen objects and improves marginally. However, the iterative explainer ($E$) can disentangle color and shape information from “\textit{gray sphere}” and “\textit{green cube},” and generalize to the unseen object “\textit{gray cude}”. During SSCR, when we intervene the counterfactual instructions to contain “\textit{gray cube},” the iterative explainer can still provide self-supervised loss to make the model consider unseen objects. Hence, SSCR can bring out obvious improvements on both F1 and RelSim, even if under the zero-shot setting.

\paragraph{Iterative Explainer \textit{vs}. Discriminator}
Fig.~\ref{fig:loss-vs} shows the learning curve of the training losses of the discriminator $D$ ($L_G$) and our iterative explainer $E$ ($L_E$). We can see that the relative decrease of $L_G$ over time is very little, which means that $D$ can barely provide extra training signal after 100 epochs. In contrast, since $E$ can supply explicit token-level loss instead of vague binary loss, $L_E$ keeps decreasing much and training the model. 
\begin{figure}[t]
    \centering
    
    \includegraphics[width=.95\linewidth]{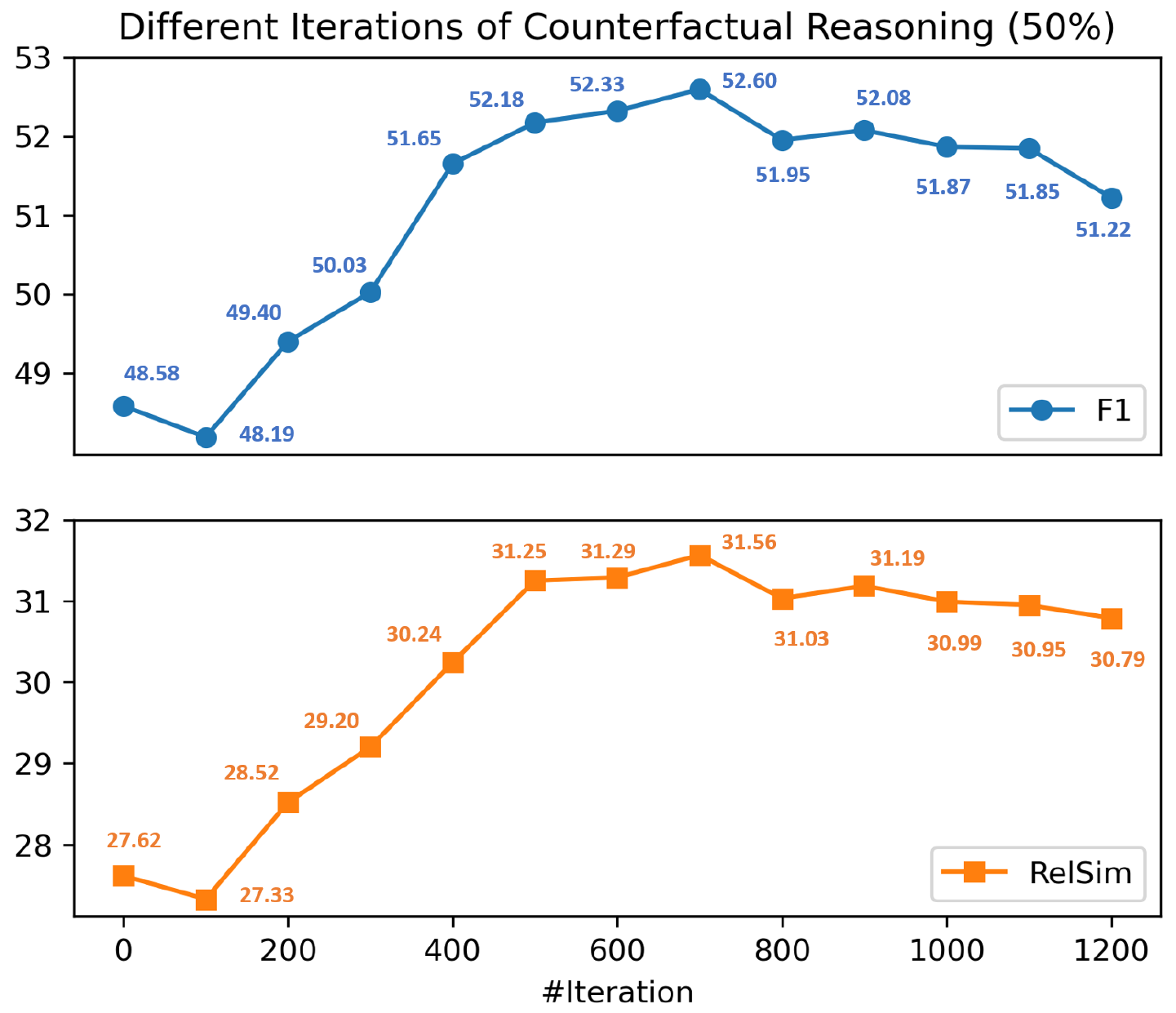}
    \vspace{-1ex}
    \caption{The validation F1 and RelSim on i-CLEVR with 50\% training data under different instructions of counterfactual reasoning.}
    
    \label{fig:sscr-iter}
    \vspace{-3ex}
\end{figure}

Table~\ref{table:sscr-vs} shows the comparison when using $D$ and our $E$ to provide the training loss during the counterfactual reasoning. If using $D$, since there are not ground-truth resulting images of those counterfactual instructions, we cannot feed them into $D$ as true examples. It can only provide training loss by discriminating predicted images as false. Therefore, using $D$ during SSCR cannot improve the model much, and may even hurt the generalizability under data scarcity, \eg, 23.9 drops to 23.0 on RelSim for 50\%. 

In comparison, since our $E$ does not suffer from data scarcity, it supports SSCR by providing meaningful training loss to perform counterfactual reasoning, and thus improves the generalizability, \eg, 23.9 increases to 32.0 on RelSim for 50\%.

\paragraph{Counterfactual Reasoning: The More The Better?}
Despite allowing the model to explore various instructions and become more generalized, excessive counterfactual reasoning may result in overfitting to existing images and degrade the performance. Fig.~\ref{fig:sscr-iter} presents the validation performance under different iterations. It shows a trade-off between the model's generalizability and the iterations of the counterfactual reasoning.
The performance keeps improving until the best 700 iteration and then drops down, possibly due to overfitting to existing images and the imperfect cost function for instruction prediction.

\begin{figure}[t]
    \centering
    
    \includegraphics[width=.7\linewidth]{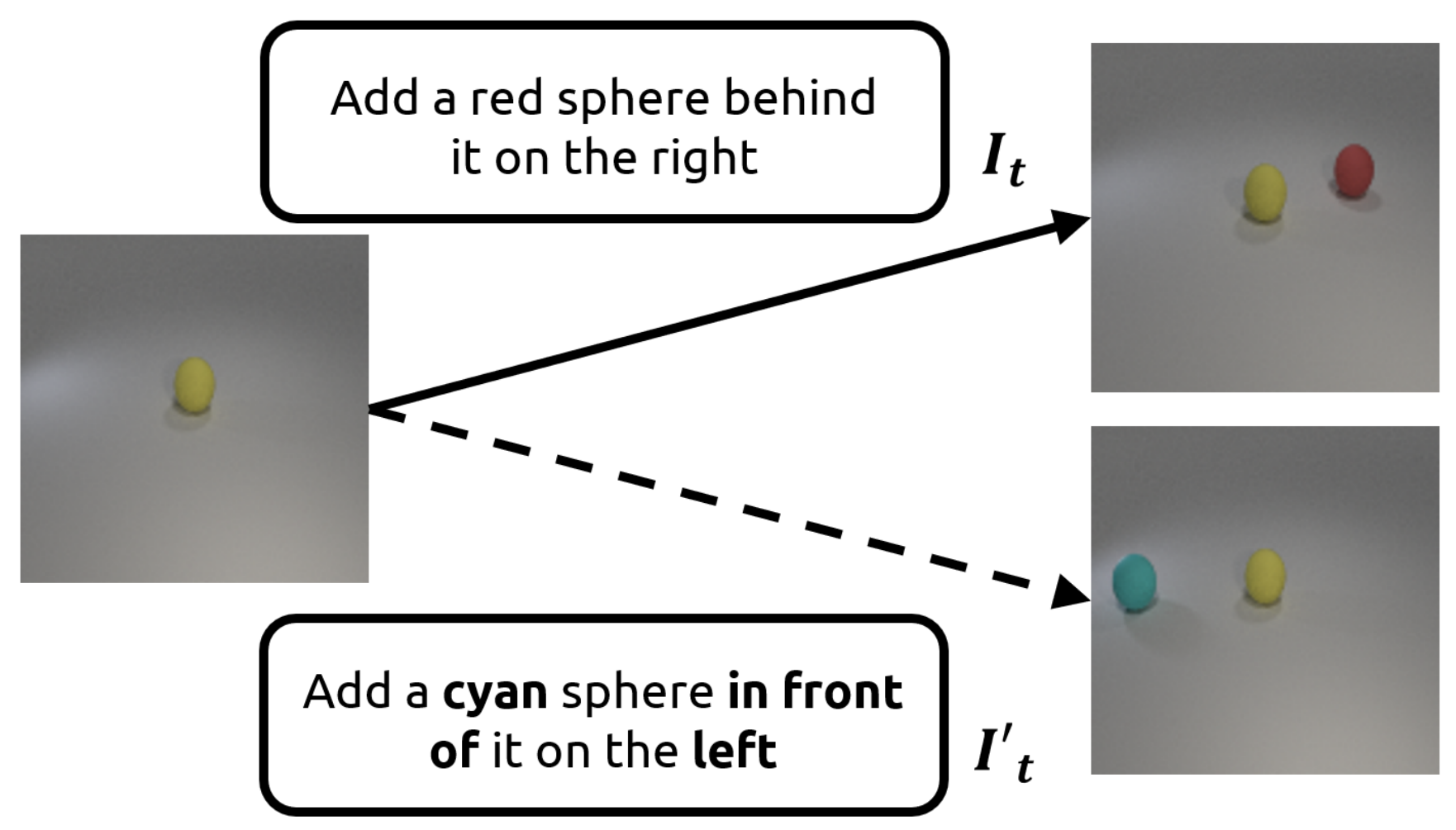}
    \vspace{-1ex}
    \caption{Example of counterfactual reasoning.}
    
    \label{fig:case-ssr}
    \vspace{-3ex}
\end{figure}

\paragraph{Qualitative Results}
Fig.~\ref{fig:case-ssr} present an example of counterfactual instructions and the predicted resulting image. We replace \textit{color token} ``red" with ``cyan"",  \textit{color token} ``behind" with ``in front of", and ``right" with "left." By considering counterfactual instructions, SSCR allows the model to explore diverse instruction-image pairs, that deals with the data scarcity issue.

Fig.~\ref{fig:case} demonstrates an example of the iterative editing on CoDraw. For baseline GeNeVA, since there is only a discriminator to provide vague loss that the pixels of those generated objects are almost broken, it makes the predicted images low quality. 
In contrast, for our SSCR, CTC can help train the generator better, which leads to defined objects. Furthermore, counterfactual reasoning also makes the predicted images more aligned to the instructions.

\section{Conclusion and Future Work}
\vspace{-1ex}
We present a self-supervised counterfactual reasoning (SSCR) framework that introduces counterfactual thinking to cope with the data scarcity limitation for iterative language-based image editing. SSCR allows the model to consider new instruction-image pairs. Despite without ground-truth resulting images, we propose cross-task consistency (CTC) to provide a more explicit training signal and train these counterfactual instructions in a self-supervised scenario. Experimental results show that our counterfactual framework not only trains the image editor better but also improves the generalizability, even under data scarcity.

For the real world, both visual and linguistic will be more complicated. To accomplish real-world image editing, large-scale pre-trained language encoders and image generators should be applied to understand the diverse instructions and model the interaction for editing. From a theoretical perspective, our SSCR is a model-agnostic framework that can incorporate with any image generator, for GAN or non-GAN architecture, to perform real-world image editing. Currently, the interactive explainer and counterfactual intervention in SSCR both improve the editing quality in the token-level. To make it more suitable for real-world images, semantic-level intervention for the diverse natural instructions can support better counterfactual reasoning. Also, a stronger explainer that explains not only token-level error but also global editing operation between two images can provide robust self-supervised loss.
\\ \\
\noindent\textbf{Acknowledgments.~}Research was sponsored by the U.S. Army Research Office and was accomplished under Contract Number W911NF-19-D-0001 for the Institute for Collaborative Biotechnologies. The views and conclusions contained in this document are those of the authors and should not be interpreted as representing the official policies, either expressed or implied, of the U.S. Government. The U.S. Government is authorized to reproduce and distribute reprints for Government purposes notwithstanding any copyright notation herein.

\bibliography{emnlp2020}
\bibliographystyle{acl_natbib}

\end{document}